\documentclass[english]{lni}

\IfFileExists{latin1.sty}{\usepackage{latin1}}{\usepackage{isolatin1}}

\usepackage{graphicx}
\usepackage{fancyhdr}
\usepackage{listings} 
\usepackage{changepage} 
\usepackage{amsmath}
\usepackage{amssymb}
\usepackage{epsfig}
\usepackage[figurename=Fig., tablename=Tab., small]{caption}[2008/04/01]
\usepackage{subcaption}

\fancypagestyle{titlepage}{
\fancyhead[RO]{\small A. Br\"omme, C. Busch, A. Dantcheva, C. Rathgeb and A. Uhl (Eds.): BIOSIG 2018, \linebreak Lecture Notes in Informatics (LNI), Gesellschaft f\"ur Informatik, Bonn 2018 \hspace{5pt} \thepage \hspace{0.05cm}} 
\fancyfoot{}}

\renewcommand{\headrulewidth}{0.4pt} 
\newcounter{savecntr}

\author{Hadi Kazemi\setcounter{savecntr}{\value{footnote}}\footnote{West Virginia University, Lane Department of Computer Science and Electrical Engineering}, \,\,\,\,\,\, Fariborz Taherkhani\setcounter{footnote}{\value{savecntr}}\footnotemark, \,\,\,\,\,\, Nasser M. Nasrabadi\setcounter{footnote}{\value{savecntr}}\footnotemark}
\title{Unsupervised Facial Geometry Learning for Sketch to Photo Synthesis}
\begin{document}

\maketitle

\renewcommand{\refname}{References}
\setcounter{footnote}{4} 
\thispagestyle{titlepage}
\pagestyle{fancy}
\fancyhead{} 
\fancyhead[RO]{\small Unsupervised Facial Geometry Learning for Sketch to Photo Synthesis \hspace{5pt} \thepage \hspace{0.05cm}}
\fancyhead[LE]{\hspace{0.05cm}\small \thepage \hspace{5pt} Hadi Kazemi et al.}
\fancyfoot{} 
\renewcommand{\headrulewidth}{0.4pt} 

\begin{abstract}
	Face sketch-photo synthesis is a critical application in law enforcement and digital entertainment industry where the goal is to learn the mapping
	between a face sketch image and its corresponding photo-realistic image. However, the limited number of paired sketch-photo training data usually prevents the current frameworks to learn a robust mapping between the geometry of sketches and their matching photo-realistic images. Consequently, in this work, we present an approach for learning to synthesize a photo-realistic image from a face sketch in an unsupervised fashion. In contrast to current unsupervised image-to-image translation techniques, our framework leverages a novel perceptual discriminator to learn the geometry of human face. Learning facial prior information empowers the network to remove the geometrical artifacts in the face sketch. We demonstrate that a simultaneous optimization of the face photo generator network, employing the proposed perceptual discriminator in combination with a texture-wise discriminator, results in a significant improvement in quality and recognition rate of the synthesized photos. We evaluate the proposed network by conducting extensive experiments on multiple baseline sketch-photo datasets.
\end{abstract}

\section{Introduction}

Automatic face sketch-photo synthesis and identification have drawn great attention because of their applications in law enforcement and digital entertainment industry \cite{wang2014comprehensive}. In law enforcement, in many cases, there is no photo of a suspect in the police database.  Therefore, a forensic or composite sketch, which is drawn by a police artist or created by a software, is the only evidence to identify the suspect. However, recognition of a suspect using a face sketch is much harder than a face photo because of the significant differences between the two modalities, such as the texture and geometric mismatching, and the sensitivity of human recognition ability to the distortions of facial features. Consequently, generating photo-realistic face images of the suspects from their forensic sketches, will significantly increase the chance of identifying them.

The idea of sketch-based photo synthesis goes back at least to Liu et al. \cite{liu2007bayesian}, who employ a probabilistic sketch-photo generation model on input-output training image pairs. Since then, several techniques have been proposed to tackle this problem, including sparse representations \cite{gao2012face}, support vector regression \cite{zhang2011face}, Bayesian tensor inference \cite{wang2009face}, embedded hidden Markov model \cite{wang2013transductive}, Markov random field model \cite{peng2016multiple}.

Recently, deep learning methods have been widely utilized in many areas \cite{taherkhani2018deep, boosari2017developing}, specifically in computer vision. With convolutional neural networks (DCNNs) becoming the common tool behind a wide variety of computer vision problems, the community has taken significant steps toward image translation tasks such as sketch-photo and photo-sketch synthesis \cite{guccluturk2016convolutional, kazemi2018facial, gatys2016image}. Peng et al. \cite{peng2017superpixel} proposed superpixel-based face sketch–photo synthesis which can estimate the inherent face structure through image segmentation \cite{zohrizadeh2018image}. Zhang et al. \cite{zhang2015end} used a six-layer convolutional neural network (CNN) to generate sketches from photos. In \cite{zhang2015end}, a new optimization objective function is utilized in the form of joint generative discriminative minimization to preserve the person's identity. To transfer image style between arbitrary images, a CNN-based framework was presented in \cite{gatys2016image} which learns generic feature representations. 

The success of DCNNs in image generation tasks is truly depends on the objective function which they are asked to minimize. A naive approach is to ask the DCNN to minimize Euclidean distance between generated image and its ground truth pixels, which tends to produce blurry results \cite{zhang2016colorful}. 
More recently, Generative Adversarial Networks (GANs) \cite{goodfellow2014generative} achieved impressive results in image generation tasks by selecting a new loss function to generate more sharp and realistic images~\cite{isola2016image, sangkloy2016scribbler, ulyanov2016texture}. The key to the success of GAN is the idea of employing an adversarial loss that encourages the generated images to be indistinguishable from real images. 

Since collecting input-output pairs for many graphics tasks, e.g., sketch-photo synthesis, is usually a laborious process, several works also tackled the image generation problem in an unpaired setting \cite{zhu2017unpaired, liu2017unsupervised, kazemi2018facial}. In sketch-photo synthesis problem, it is even more difficult to train the network in a supervised fashion, which requires the corresponding pair of images from both the source (sketch), and the target (photo-realistic image) domains. There are two main reasons for this difficulty: First, since each artist or software has a distinct painting style, the network usually needs fine tuning on very limited number of unseen sketch styles. Second, because of the geometric mismatching between the sketches and their ground truth photos, it is almost impossible to learn a single mapping between the two domains. Consequently, in this work, we aim to design a framework which can be trained in an unpaired setting.

Although our work is based on CycleGAN~\cite{zhu2017unpaired}, one of the most successful unsupervised image-to-image translation frameworks, it differs from that in two effective ways. First, the CycleGAN, similar to other unsupervised GAN frameworks, is based on a \textit{cycle consistency} constraint, which indicates that we should be able to reconstruct the input image from the synthesized photo. This constraint is defined as Euclidean distance between the pixels of the input and its reconstruction. However, when we are dealing with images of different modalities, e.g., sketch-photo synthesis problem, this definition can decrease the flexibility of the network or even prevents it from convergence \cite{kazemi2018facial}. This problem caused by the excessive force on the network to keep information in the pixel level. To overcome this issue, we propose to use the perceptual-loss~\cite{johnson2016perceptual} as an alternative definition of the cycle consistency. Using perceptual-loss, the network is required to only keep the high-level facial features in the reconstructed photo which matter in terms of face identification.

The second, and the main, difference between our framework and the CycleGAN is employing a new discriminator, referred to as geometry-discriminator, to learn facial geometry. Isola et al.~\cite{isola2016image} confirmed that the discriminator of CycleGAN, namely PatchGAN~\cite{isola2016image}, only captures local information like the textural information. Since, in statistical point of view, GAN learns the distribution of the target domain, speaking of our application, PatchGAN can only force the network to learn facial texture prior. The learned texture prior is then utilized to translate an input face sketch into a photo-realistic face while keeping the exact facial geometry of the input sketch. In contrast, in this work, our goal is to learn the facial geometry prior as well, which encourages the network to learn generating photos that are realistic in terms of the texture and geometry of the face. To this end, the aforementioned geometry-discriminator, is trained to distinguish between the real and synthesized photos by their high-level facial features. The experimental results show how learning the facial geometry help the network to generate realistic face photos. 

\section{Generative Adversarial Networks (GANs)}
\textbf{Supervised Setting:}
GANs \cite{goodfellow2014generative} are a group of generative models which learn the statistical distribution of training data, allowing us to synthesize data samples by mapping a random noise $z$ to an output image $y$: $G(z): z \longrightarrow y$, where $G$ is the generator network. GAN in its conditional setting (cGAN) is proposed in \cite{isola2016image} which learns a mapping from an input $x$ and a random noise $z$ to the output image $y$: $G(x, z): \{x, z\} \longrightarrow y$, using an autoencoder network. The generator model, $G(x, z)$, is trained to generate an image which is not distinguishable from "real" samples by a discriminator network, $D$. Simultaneously, the discriminator is learning, adversarially, to discriminate between the "fake" generated images by the generator and the real samples from the training dataset. 
Therefore, the objective of cGAN can be expressed as:
\begin{align} \label{eq:ad_loss}
l_{GAN}(G,D) =  \mathbf{E}_{x,y\sim p_{data}}[\log D(x,y)]+  \mathbf{E}_{x, z\sim p_{z}}[\log (1 - D(x,G(x, z)))],
\end{align}
where $G$ attempts to minimize it and $D$ tries to maximize it. Since the adversarial loss is not enough to guarantee that the trained network generates the desired output, one may add an extra Euclidean distance term to the objective function to generate images which are near the ground truth. Consequently, the final objective is defined as follows:
\begin{align} \label{eq:cgan1}
G^* = \arg \min_G \max_D l_{GAN}(G,D) + \lambda l_{L1}(G),
\end{align}
where $\lambda$ is a weighting factor and $l_{L1}(G) = \parallel y - G(x,z) \parallel_1$.

\textbf{Unsupervised Setting:}
GAN in its unsupervised setting aim to capture a shared representation between the images in two different domains, all in the absence of any paired training data samples. CycleGAN \cite{zhu2017unpaired} is one of the most successful works that addressed this problem.  Their model includes two generators; the first one maps $x$ to $y$: $G_y(x): x \longrightarrow y$ and the other does the inverse mapping $y$ to $x$: $G_x(y): y \longrightarrow x$. There are two adversarial discriminators $D_x$ and $D_y$, one for each generator. Unsupervised GAN frameworks, including the CycleGAN, owe their success to a cycle-consistency constraint assumption, which indicates that if a source image is mapped to an image in the target domain, it can be mapped back to the original image in the source domain. To this end, the CycleGAN minimizes the following loss function:
\begin{align}
l_{cyc}(G_x, G_y) =  \mathbf{E}_{x\sim p_{data}[\parallel x - G_x(G_y(x)) \parallel_1]} + \mathbf{E}_{y\sim p_{data}[\parallel y - G_y(G_x(y)) \parallel_1]}.
\end{align}
Taken together, the full objective function of CycleGAN is
\begin{align}\label{eq:cgan_loss}
l_(G_x,G_y,D_x,D_y)=l_{GAN}(G_x,D_x)+l_{GAN}(G_y,D_y)+\lambda l_{cyc}(G_x, G_y),
\end{align}
where $\lambda$ is a weighting factor to control the importance of the objectives and the whole model is trained as follows
\begin{align} \label{eq:cgan}
G_x^*, G_y^* = \arg \min_{G_x, G_y} \max_{D_x, D_y} l(G_x,G_y,D_x,D_y).
\end{align}

\textbf{Limitations of CycleGAN:}
The CycleGAN model, and some other similar models, have two fundamental issues: first, they have defined cycle consistency as a Euclidean distance between the pixels of the input image and its reconstruction. From our experiments, when the domains differ substantially in complexity, this definition of cycle consistency causes a degradation in quality of synthesized images, and even can destabilize the training process. Second, the discriminator of the CycleGAN is PatchGAN~\cite{isola2016image} that only penalizes structure at the scale of patches, and can be considered as a form of texture/style loss. Consequently, it is unable to learn the distribution of facial geometry and sticks to the geometry of input sketch for inference. Next, we discuss how to extend the CycleGAN, enabling it to generate realistic face images, in terms of texture and geometry. 

\section{Framework}
\textbf{Cycle-Consistency:}
To learn complex cross-domain relationships, we define the Euclidean distance on high-level feature space as the cycle consistency, which is known as \textit{perceptual loss} \cite{johnson2016perceptual}. Similar to \cite{johnson2016perceptual}, we make use of a VGG-16 network, $\Phi$, pretrained for face verification on CMU Multi-PIE \cite{gross2010multi} dataset, as a fixed loss network. This loss network defines a feature reconstruction loss that measure differences in high-level content between images. Let $\Phi_j(x)$ denote the feature maps of the $j^{th}$ layer of the loss network for the input image $x$. Then the $j^{th}$ layer perceptual loss between images $x$ and $y$ is defined as
\begin{align} \label{eq:percept}
l_{per}^j(x,y) = \frac{1}{N_j}\parallel \Phi_j(x) - \Phi_j(y) \parallel_2^2,
\end{align}
where $N_j$ is the number of perceptrons in the $j$th layer. In this work, we defined the cycle-consistency loss as the average perceptual loss in multiple layers of the loss network as: 
\begin{align} \label{eq:cycle2}
l^p_{cyc}(G(x)) = \frac{1}{3}\mathbf{E}_{x\sim p_{data}} \big [l_{per}^7(x,\hat x) + l_{per}^{10}(x,\hat x) + l_{per}^{13}(x,\hat x) \big ],
\end{align}
where $\hat x = G_x(G_y(x))$.

\textbf{Geometry-Discriminator:}
\begin{figure}[t]
	\begin{center}
		\includegraphics[width=0.95\linewidth]{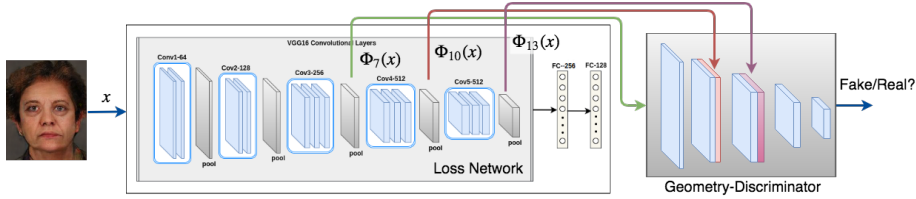}
	\end{center}
	\caption{Geometry-discriminator, $D^g$.}
	\label{fig:dis}
\end{figure}
A straightforward solution for extending the CycleGAN to learn the facial geometry is to equip it with a geometry-discriminator $D^g$. In contrast to PatchGAN, the input to this discriminator, is the feature maps of the loss network, which is shared with the perceptual cycle-consistency. In other words, $D^g$ tries to distinguish between the synthetic and real images from their high-level features, which represent the geometry of the face and its components. We use feature maps in the same level as perceptual loss, i.e., layers 7, 10, and 13. Since the spatial sizes of the feature maps in different layers of the loss network are different, we cannot input them to $D^g$ at the same time. Alternatively, we input the feature map with the largest size, i.e., $7^{th}$ layer, to the discriminator. Since all the convolutional layers of $D^g$ have a stride 2, the spatial size of each layer's output is half of its input size. Therefore, we can concatenate the feature maps of the $10^{th}$ layer of loss network with the output of the first convolutional layer of $D^g$. Similarly, we can concatenate the feature maps of the $13^{th}$ layer of loss network with the output of the second convolutional layer of the geometry-discriminator. Figure~\ref{fig:dis} visualizes this process. Since our images are in the size of 256$\times$256 pixels, the spatial size of feature maps in the $7^{th}$ layer of the loss network is of 32$\times$32. Note that the geometry-discriminator needs 5 convolutional layers to output a single prediction for a given image.

\begin{figure}[t]
	\begin{center}
		\includegraphics[width=0.95\linewidth]{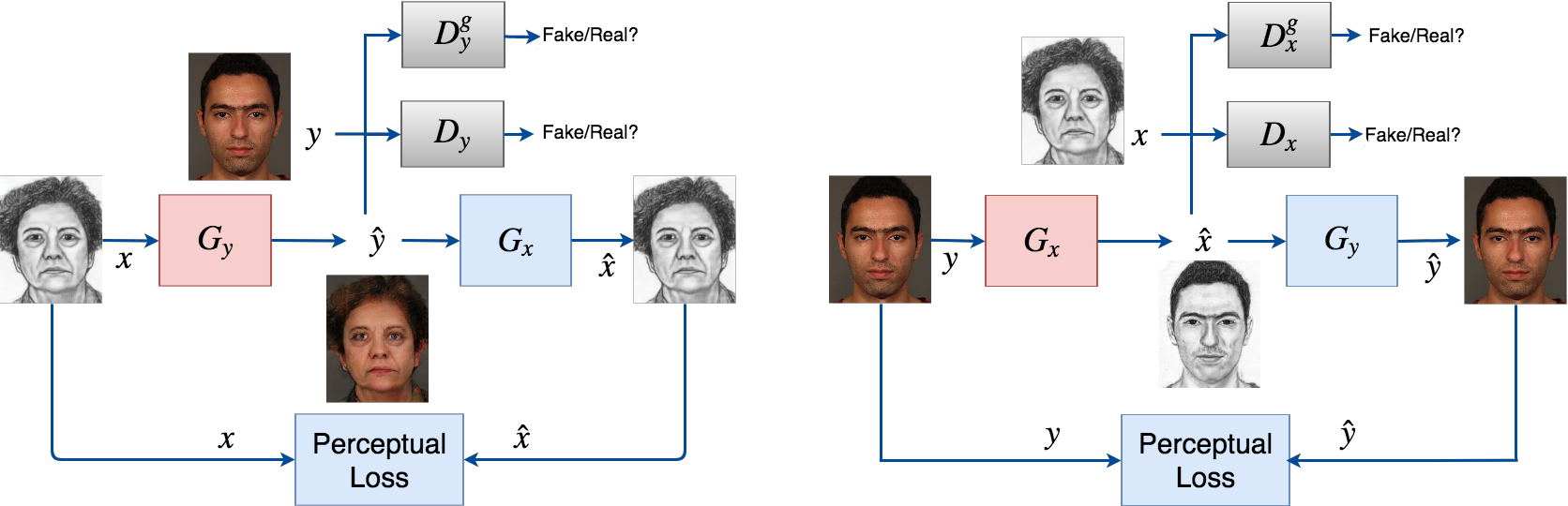}
	\end{center}
	\caption{Our framework shared generator and PatchGAN discriminator architectures with the CycleGAN, while we replaced the pixel-wise cycle-consistency with perceptual loss, and added a geometry-discriminator, $D^g$, to the model.}
	\label{fig:cycle1}
\end{figure}

\textbf{Full Objective Function:}
Similar to CycleGAN we train the network in two cycles of $X \rightarrow Y \rightarrow X$, and $Y \rightarrow X \rightarrow Y$, where $X$ represents the sketch domain, and $Y$ stands for the photo-realistic domain. Figure~\ref{fig:cycle1} illustrates the overall schematic of the framework. Taken together, the full objective function of our frameworks is defined as follows:
\begin{align}\label{eq:our_loss}
l_(G_x,G_y,D_x,D_y)&=l_{GAN}(G_x,D_x)+l_{GAN}(G_x,D^g_x)+l_{GAN}(G_y,D_y)+l_{GAN}(G_y,D^g_y)\\ \nonumber &+\lambda_1 l^p_{cyc}(G_x)+\lambda_1 l^p_{cyc}(G_y),
\end{align}
where  the  hyper-parameters $\lambda_1$ and $\lambda_2$ control the weights of the objective terms.

\section{Experiments}
\textbf{Datasets:} 
The e-PRIP dataset \cite{mittal2017composite}, contains the total of 123 identities, which we partitioned into a training set of 100 identities, and a testing set of 23 identities. We made use of its composite sketches created by the Identi-Kit and the FACES tools. For more evaluation, we used CUHK Face Sketch FERET Database (CUFSF) \cite{phillips2000feret} with the total of 1194 identities (1000 identities for training and the remaining 194 identities for testing). Since CUFSF photos are in grayscle, we used the FERET color dataset as photo domain to synthesize color photos.

\textbf{Evaluation:} We evaluate the following three models: Our proposed model with the perceptual cycle-consistency but with (w/ $D^g$) and without (w/o $D^g$) geometry-discriminator, and the CycleGAN as the baseline.

\subsection{Qualitative Evaluation}
\begin{figure}[htbp]
	\centering
	\begin{minipage}{.48\textwidth}
		\centering
		\includegraphics[width=.95\linewidth]{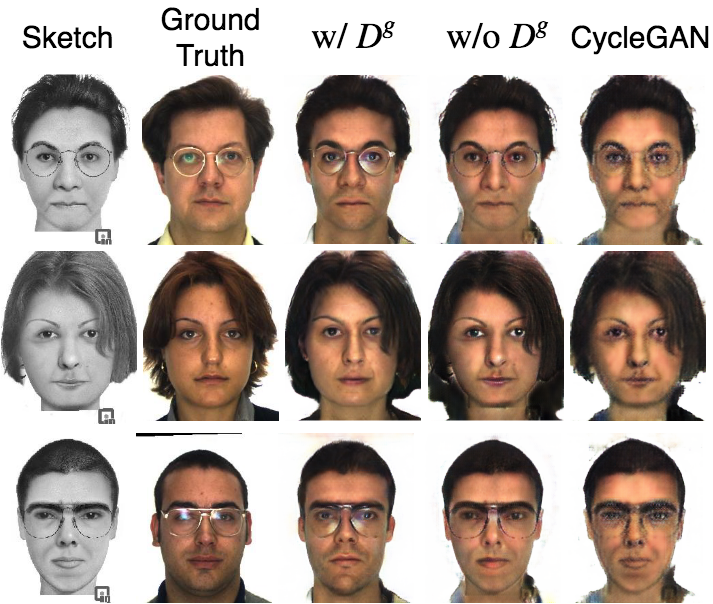}\\
		(a) FACES PRIP Dataset
	\end{minipage}\hfill
	\begin{minipage}{.48\textwidth}
		\centering
		\includegraphics[width=.95\linewidth]{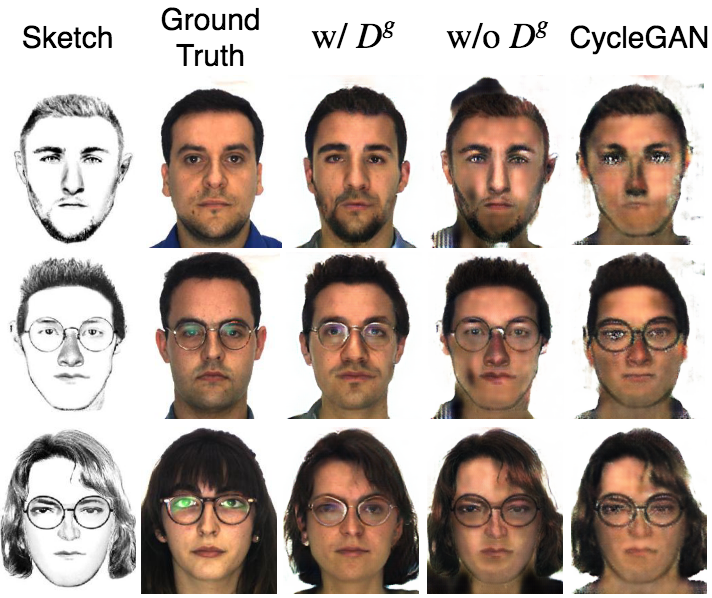}\\
		(b) Identi-Kit PRIP Dataset
	\end{minipage}
	\includegraphics[width=.95\linewidth]{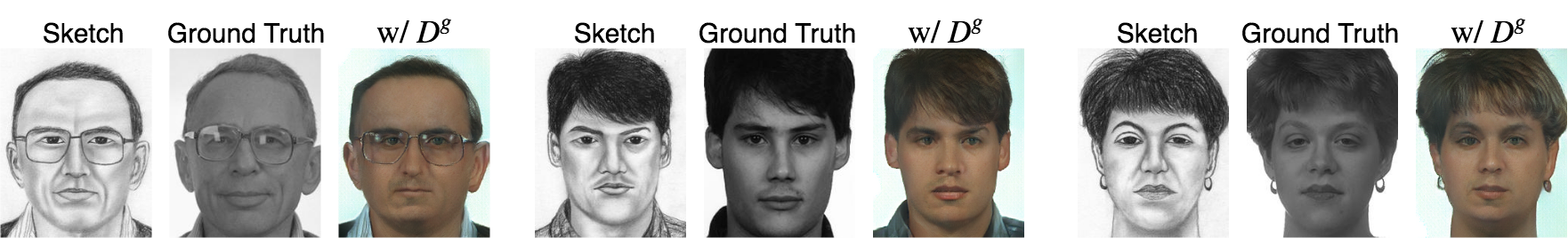}\\
	(c) CUFSF Dataset
	\caption{From left to right: Input sketch, ground truth photo, our framework with geometry-discriminator, our framework without geometry-discriminator, and the CycleGAN.}
	\label{fig:comparison}
\end{figure}

Figure~\ref{fig:comparison} shows the qualitative comparison results on the EPRIP Identi-Kit and FACES datasets, and more evaluation on CUFSF dataset. The results are reported after training for 200 epochs using the Adam optimizer with learning rate of 0.0002. The CycleGAN almost acts as coloring tool and cannot generate realistic sharp textures. The quality of textures, however, significantly improves by replacing the pixel-wise cycle-consistency with perceptual loss. That empowers the network to generate more sharp images by relieving the constraint to keep the information in pixel-level. Equipping the model with geometry-discriminator, we observe results which are realistic both texture-wise and geometry-wise.

\subsection{Quantitative Evaluation}

\begin{table}[htb]
	\centering
	\begin{tabular}{|l|c|c|c|c|c|c|}
		\hline
		& \multicolumn{2}{c|}{Semantic Accuracy} & \multicolumn{2}{c|}{Fooling Rate (\%)} & \multicolumn{2}{c|}{Verification Accuracy (\%)} \\ \hline
		\multicolumn{1}{|c|}{Method} & FACES           & Identi-Kit          & FACES        & Identi-Kit        & FACES             & Identi-Kit             \\ \hline
		Ours w/ $D^g$                & 0.173               & 0.179                    & 42.3            & 40.1       & 62.3 $\pm$ 2.0    & 60.8 $\pm$ 3.0      \\ \hline
		Ours w/o $D^g$               & 0.207               & 0.231                    & 12.8            & 10.4       & 58.0 $\pm$ 2.7    & 56.3 $\pm$ 2.8      \\ \hline
		CycleGAN                     & 0.263               & 0.284                    & 6.3            & 5.2         & 55.7 $\pm$ 2.6    & 52.4 $\pm$ 3.2      \\ \hline
	\end{tabular}
	\caption{The quantitative comparison of our framework with the CycleGAN on the EPRIP dataset.}
	\label{table:results}
\end{table}
We show the average distance of generated output images and their ground-truth images, in deep feature space, over a test set of 23 samples in Table~\ref{table:results}. The results clearly confirm that the proposed technique substantially improves the semantic accuracy of the synthesized images while stays faithful to the input sketch. Since the primary reason for sketch-photo synthesis is distributing the realistic synthesized images to the public for suspect identification, we conducted two other experiments, visual realism based on human judgments, and face verification accuracy using a pre-trained face verifier. For visual realism test, we successively display a real and generated image, in a random order, to multiple subjects for 1 second each and ask them to identify the fake, and then we calculate the "fooling" rate. The fooling rates of the three methods are also reported in Table~\ref{table:results}. 

Finally, to investigate the effect of learning facial geometry on face verification accuracy, we utilized a VGG16 face verifier pre-trained on the CMU Multi-PIE dataset \cite{gross2010multi}. For each sketch, the results of different methods are used for face verification against the entire test gallery. Next, the verification accuracy of each method is calculated and reported in Table~\ref{table:results}. Note that we used all 123 identities in the photo gallery. The evaluation process is repeated 10 times with a random selection of training and testing sets. The verification accuracy confirms the effectiveness of learning the facial geometry for artifacts removal and identification performance improvement. Generating sharp images and comparing them with the CycleGAN may have also contributed to our framework's higher accuracy.

\section{Conclusion}
The qualitative and quantitative results of this paper suggest that cycle-consistency and discrimination using high-level perceptual features are promising approaches for improving the quality of translated images in terms of facial texture and geometry. These features are extracted from a pre-trained face verifier network which let to train the sketch-photo synthesis framework on small datasets. 

\bibliographystyle{lnig}
\bibliography{egbib}

\end{document}